\documentclass[twocolumn,conference,letterpaper]{IEEEtran}
\usepackage[left=0.75in,right=0.75in,top=1in,bottom=0.75in]{geometry}

\usepackage{cite}
\usepackage{amssymb,amsmath,latexsym,amsfonts,amsthm,mathtools}

\usepackage{algorithmic}
\usepackage{graphicx}
\usepackage{float}
\usepackage{subcaption}
\usepackage{textcomp}
\usepackage{xcolor}
\definecolor{darkpastelpurple}{rgb}{0.59, 0.44, 0.84}
\usepackage{hyperref}
\hypersetup{colorlinks, breaklinks, citecolor=blue, linkcolor=blue, urlcolor=blue}
\usepackage[nameinlink]{cleveref}
\Crefformat{figure}{#2Fig.~#1#3}
\Crefmultiformat{figure}{Figs.~#2#1#3}{ and~#2#1#3}{, #2#1#3}{ and~#2#1#3}

\usepackage{tikz}
\usetikzlibrary{automata, shapes, arrows, calc, arrows.meta, fit, positioning}

\theoremstyle{plain}

\newtheorem*{problem*}{Problem}
\theoremstyle{remark}

\theoremstyle{definition}

\usepackage{mathrsfs}

\usepackage{newtxmath}
\usepackage{booktabs}

\IEEEoverridecommandlockouts

\begin{document}
	\title{A Study of Feature Selection and Extraction Algorithms for Cancer Subtype Prediction}
	\author{Vaibhav Sinha, Siladitya Dash, Nazma Naskar, and Sk Md Mosaddek Hossain
		\thanks{V. Sinha, S. Dash, and N. Naskar are with the School of Computer Engineering, Kalinga Institute of Industrial Technology, Bhubaneswar, India- 751024. (e-mails: vaibhavsinha.cs@gmail.com, siladityadash.cs@gmail.com, nazma.naskarfcs@kiit.ac.in).\newline Sk Md M. Hossain is with the Department of Computer Science and Engineering, Aliah University, Kolkata, India-700156. (e-mail: mosaddek.hossain@aliah.ac.in).}	
	}
	
	\maketitle
	\thispagestyle{empty}
	
	\begin{abstract}


		In this work, we study and analyze different feature selection algorithms that can be used to classify cancer subtypes in case of highly varying high-dimensional data. We apply three different feature selection methods on five different types of cancers having two separate omics each. We show that the existing feature selection methods are computationally expensive when applied individually. Instead, we apply these algorithms sequentially which helps in lowering the computational cost and improving the predictive performance. We further show that reducing the number of features using some dimension reduction techniques can improve the performance of machine learning models in some cases. We support our findings through comprehensive data analysis and visualization.
	\end{abstract}
	
	\vspace{0.3cm}
	
	
	\section{Introduction}\label{sec:introduction}
	Cancer subtype prediction has evidently been of great importance for guiding the treatment of patients suffering from any type of cancer. For several years, researchers and biologists have been studying how subtype identification can be used to plan treatments and diagnosis \cite{10.1007/978-3-030-34139-8_39,10.3389/fgene.2019.00020} . Identifying potential bio-markers in a patients body by studying the gene expression values has been a subject of research for a long time \cite{biomarkers}. With the advancement of machine learning, it is possible to classify cancer subtypes and study the results in an elegant manner. However, since human cells contain thousands of genes, all of which may or may not play a crucial role in the identification of a cancer or its subtypes, the need to filter the genes using various parameters and metrics arises.
	
	The study in \cite{10.1007/978-3-030-34139-8_39} has shown how a feature selection approach can impact the performance of the supervised predictive model when identifying cancer subtypes. In \cite{10.1007/978-3-030-34139-8_39}, the authors used a kernel-based clustering method for gene selection, and selected genes based on their weights. The work in \cite{DBLP:conf/icml/YuL03} proposed a correlation-based filter solution for feature selection in high-dimensional data by introducing a novel concept of predominant correlation. This method also identified redundancy among relevant features. A novel approach to combine feature selection and transductive support vector machine (TSVM) was proposed in \cite{6334430}  which resulted in improved accuracy as compared to standard SVM algorithms.  In \cite{doi:10.1080/03772063.2021.1878062}, the authors proposed an unsupervised feature selection method for cancer prediction. They used a Singular Value Decomposition (SVD) Entropy method for feature ranking. A novel unsupervised Similarity Kernel Fusion (SKF) was proposed in \cite{10.3389/fgene.2019.00020} which made use of spectral clustering on the integrated kernel to predict cancer subtypes. 
	
	There have been studies using deep learning for cancer subtype classification as well. A novel supervised cancer classification framework, Deep Cancer subtype Classification (DeepCC) was proposed in \cite{Gao2019DeepCCAN} which was based on deep learning of functional spectra quantifying activities of biological pathways. A stacking ensemble based deep learning approach was also proposed in \cite{PMID:34341396} which was based on One-dimensional Convolutional Neural Network (1D-CNN) and Least Absolute Shrinkage and Selection Operator (LASSO) as the feature selection methods. The study in \cite{Mostavi2020} presented novel CNN models to predict cancer types based on gene expression profiles and elucidated biological relevance of cancer marker genes. The authors in \cite{ijerph18042197} used deep autoencoder for feature extraction and used an oversampling algorithm to handle data imbalance. Such approaches demonstrate how effective feature selection can be when dealing with high-dimensional data for classification. 
	
	However, most of these approaches were applied on a particular type of cancer and omics. In order to understand and compare the effects of common feature selection algorithms for predicting cancer subtypes, we aim to present a comparative study of how the effectiveness of these algorithms vary when tested against different omics of various cancer types.  It is important to note that a majority of these aforementioned works used a single feature selection algorithm. Also, in case of data with high dimension and low sample size, deep learning approaches may find it difficult to learn relevant patterns, and in some cases, overfit the training data. Such approaches also take a significant amount of time to produce results. 
	
	To overcome these shortcomings, along with improving the performance of a relatively simpler machine learning model, we use three different feature selection techniques sequentially. When these methods are used individually on the complete training data, the estimator takes a significant amount of time going over every single feature to calculate its importance score. One of these methods uses k-fold cross validation to calculate the optimal number of features. This makes the methods computationally expensive and the resultant number of features is still quite high. Using these methods sequentially provides only the relevant features to the next algorithm in the sequence which in turn reduces the computation time and the dimension of the data. 
	
	The remainder of this paper is organized as follows. After a brief introduction in \Cref{sec:introduction}, data processing and workflow are presented in \Cref{sec:dataprocessing}. Machine learning modeling of the relevant data is presented in \Cref{sec:modeling}, followed by results and analysis in \Cref{sec:results}. Finally, \Cref{sec:conclusions} presents concluding remarks and discusses some outlines for future work.

	\section{Data Processing}\label{sec:dataprocessing}
\begin{figure}[!h]
		\centering
		\includegraphics[width=1\linewidth]{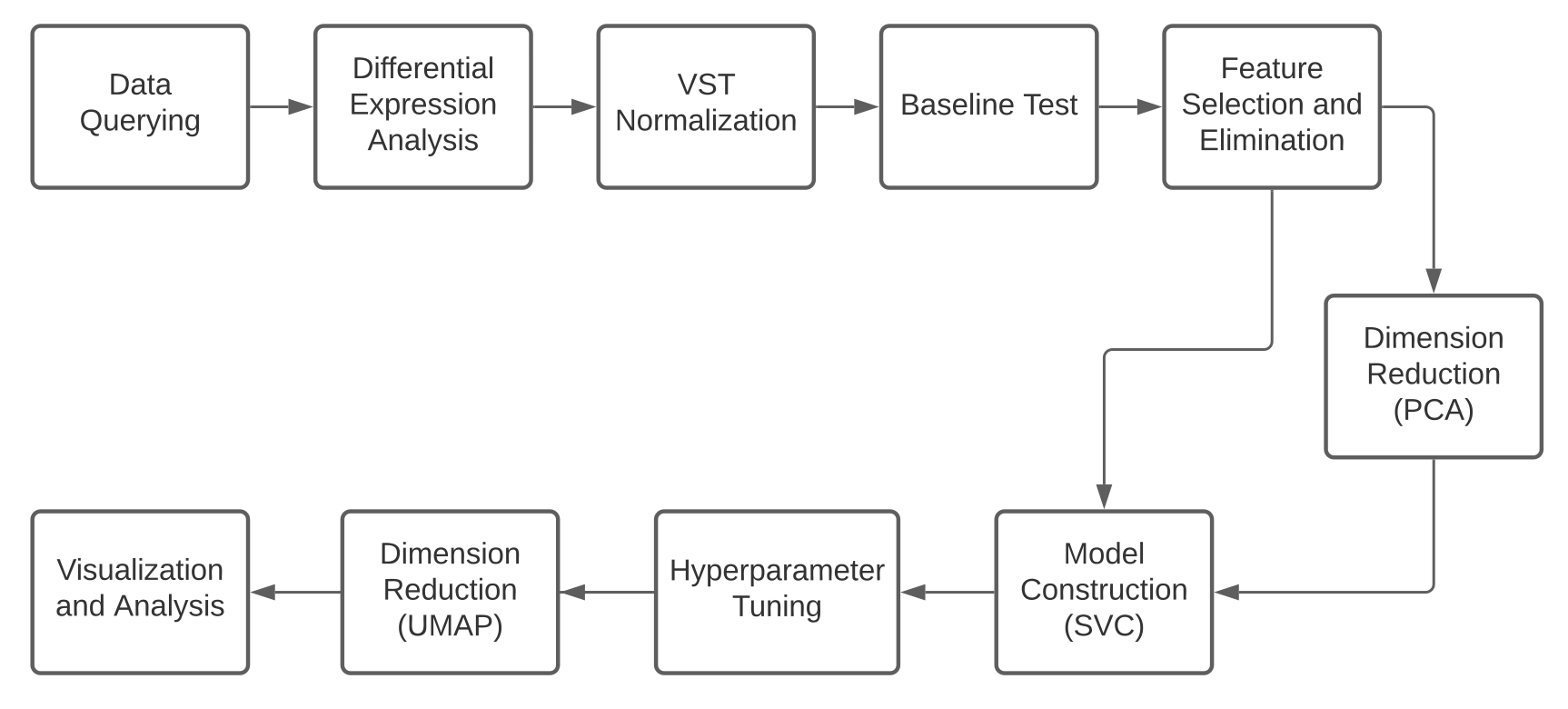}
		\caption{Workflow of Cancer Subtype Prediction.}
		\label{fig:workflow}
	\end{figure}

	We use various classification algorithms to compare the effect that the individual feature selection methods have on the omics data. Since omics data sets have a very large number of features, ranging from 1,000 to 50,000, it is necessary to perform various feature selection methods, and dimensionality reduction algorithms to make the data sets more suitable for predictive modeling and analysis. To this aim, we perform normalization of the gene expression values to scale the values of the data set. After preprocessing the values, we use various machine learning models to make cancer subtype predictions. By using advanced dimensionality reduction techniques like Uniform Manifold Approximation and Projection (UMAP), we further analyze the data for additional insights and inferences. A detailed overview of the entire process is  depicted in \Cref{fig:workflow}, and discussed next.	
	
	\subsection{Dataset}
	Cancer subtypes are the smaller groups that a particular type of cancer is divided into based on specific characteristics of the cancer cells. It is essential to know the subtype of cancer to plan treatment and determine prognosis. The National Institutes of Health (NIH) created The Cancer Genome Atlas (TCGA) program to obtain a comprehensive understanding of the genomic alterations that underlie all major cancers. TCGA provides access to various omics data like gene expression, DNA methylation, copy number variations, etc. across different cancer types. Such data are highly variable and high-dimensional. Since a particular cancer can have multiple subtypes, their identification is essential for providing patients with the necessary treatment.
	
	In the interest of our analysis, we download ten datasets from the Genomic Data Commons (GDC) Data Portal which includes five cancer types having two omics each. The two omics are miRNA Expression and Gene Expression (RNASeq), and the five cancer types are Head-Neck Squamous Cell Carcinoma (TCGA-HNSC), Kidney Renal Clear Cell Carcinoma (TCGA-KIRC), Kidney Renal Papillary Cell Carcinoma (TCGA-KIRP), Lung Adenocarcinoma (TCGA-LUAD), and Lung Squamous Cell Carcinoma (TCGA-LUSC). We use the harmonized versions of all the datasets for our study.
	
	\subsection{Required Tools}
	In our work, we use the R programming language for initial preprocessing and analysis of our data. We use TCGAbiolinks package \cite{tcgabiolinks,tcgaworkflow,gtex} for downloading, preparing, and performing some preprocessing tasks on the data. The initial analysis of data also requires the EDASeq package \cite{edaseq}. We use the SummarizedExperiment package \cite{assays} to use its container which contains one or more assays. The assays are represented by matrix-like objects. The rows typically represent genomic ranges of interest and the columns represent samples. We use the DESeq2 package \cite{deseq2} for further analysis of RNASeq data using DESeqDataSet object and for normalizing the data. For Differential Expression Analysis (DEA), we use the edgeR package \cite{edger}. After getting the normalized data, we use Python programming language for performing feature selection and extraction, and machine learning modeling.
	
	\subsection{Preparing Data}
	On a predictive modeling problem, machine learning algorithms are used to map the input variables to a target label. However, we cannot fit these algorithms on raw data. We need to transform the data into such a representation that we can best explore its underlying structure and use a suitable algorithm for getting the satisfactory performance out of the machine learning model with the given resources.
	
	
	We search the GDC database for parameters like data category, data type, workflow type, and file type to access the data. For RNASeq, we select the data category as Transcriptome Profiling, the data type as Gene Expression Quantification, and the workflow type as HTSeq-Counts. For miRNA, we select the data category as Transcriptome Profiling, and the data type as miRNA Expression Quantification. 	After getting the queried data, we take all the samples and select only the Primary Solid Tumor (PST) and Solid Tissue Normal (STN) samples. We perform down-sampling and add clinical information, TCGA molecular information, and respective cancer subtype information to the samples.
	
	\subsection{Differential Expression Analysis (DEA)} 
	DEA involves taking the normalized read count data and performing statistical analysis to discover quantitative changes in expression levels between experimental groups. For example, statistical testing can be used to decide whether, for a given gene, an observed difference in read counts is significant, that is, whether it is greater than what would be expected just due to natural random variation. The goal of DEA is to determine which genes are expressed at different levels between conditions. These genes can offer biological insight into the processes affected by the condition(s) of interest. The count data used for DEA represents the number of sequence reads that originated from a particular gene. The higher the number of counts, the more reads associated with that gene, and the assumption that there was a higher level of expression of that gene in the sample.  Before performing DEA on the dataset, we preprocess the data for removing samples (using the Pearson Correlation Coefficient) with low correlation that were possible outliers.
	
	In case of RNASeq count data, major technology-related artifacts and biases affect the expression measurements and therefore normalizing the expression values based on some metric is an important aspect before performing the DEA. The GC-content bias is one such bias which describes the dependence between fragment count and GC-content found in Illumina sequencing data. This bias can dominate the signal of interest for analyses that focus on measuring fragment abundance within a genome, such as RNASeq \cite{10.1093/nar/gks001}. We use \emph{TCGAanalyze\_Normalization} which uses within-lane normalization procedures to adjust for GC-content effect (or other gene-level effects) on read counts by using Loess robust local regression, global-scaling, and full-quantile normalization \cite{edaseq}, and between-lane normalization procedures to adjust for distributional differences between lanes (e.g., sequencing depth) by using global-scaling and full-quantile normalization \cite{Bullard2009EvaluationOS}.
	
	After normalizing the expression values, we use a filtration process to remove all the genes except the ones which have quantile mean values higher than the threshold defined across all samples. Next, we perform DEA on the remaining gene expression values where a negative binomial generalized log-linear model is fit to the read counts for each gene that outputs another set of filtered genes, thereby  clearing the cutoff set for the DEA procedure.
	
	\subsection{Data Normalization}
	In gene expression data analysis, normalization is used to correct the measurement errors and bias introduced in the acquisition of data. The errors and bias may be introduced due to many factors such as concentration of target RNA sequence, instrumental noise, etc.	From the data obtained after performing DEA, we create a \emph{DESeqDataSet} object to store the read counts and the intermediate estimated quantities during statistical analysis. Then, we handle the under-expressed genes by removing the rows which have a sum of less than a certain threshold (here, it is selected as 10). On the resulting data, we calculate Variance Stabilizing Transformation (VST) from the fitted dispersion-mean relation(s) and transform the count data (normalized by division by the size factors or normalization factors), yielding a matrix of values which are approximately homoskedastic (having constant variance along the range of mean values).
	
	\subsection{Feature Selection and Dimension Reduction}
	Reducing the number of input variables can reduce the computational cost, improve the performance of the model, and render the underlying algorithm fast and effective. However, the choice of the algorithm depends on the type of data. Problems involving gene expression can be high-dimensional in nature, i.e., they can contain tens of thousands of features. With staggeringly high number of features, the calculations become extremely difficult. For better performance of the machine learning models, we aim to reduce the dimensions of the data.
	
	In our work, we use the python library scikit-learn \cite{scikit-learn} for feature selection, dimensionality reduction, and machine learning modeling. First, we perform tree-based feature selection using random forest classifier. This is used to compute impurity-based feature importance, which in turn are used to discard irrelevant features. We then perform Recursive Feature Elimination with Cross-Validation (RFECV) on the obtained relevant features. We use linear kernel SVC as the estimator to find the optimal number of features based on the accuracy score on the training set. We perform stratified two-fold cross-validation to reduce bias. Finally, we apply Recursive Feature Elimination (RFE) with the optimal number of features using the same estimator as before. RFE selects features by recursively considering smaller and smaller sets of features. We train the estimator on the initial set of features and obtain their importance score and prune the least important features.  We select the optimal number of features by recursively repeating this step.
	
	\section{Machine Learning Modeling}\label{sec:modeling}
	During the modeling phase, we perform a baseline test on various models and select the models that perform relatively better. We then try to  improve the performance by tuning various hyper-parameters and introducing various augmentations to the models. We perform the baseline test on six models, namely Decision Tree Classifier, Random Forest Classifier, KNeighbors Classifier, SVC, Gaussian Naive Bayes, and Logistic Regression. 
	
	We use the aforementioned feature selection algorithms  to filter out features with low importance in the prediction process of the model. In the tree-based feature selection, we use a Random Forest classifier to select the features with high importance score, which are then used to make the prediction. For RFE, we use SVC as the estimator to recursively select features until the required amount of features are reached. Using all these algorithms separately does improve the result, with the recursive techniques giving comparatively better results. But the recursive techniques are computationally very expensive as well. This can be attributed to the fact that recursive algorithms are generally slower by design and the number of features being very high, it becomes an even more time consuming process. However, the tree-based method is comparatively faster, while stripping the dataset of a decent number of features. Each method filters out the features based on different parameters and metrics. What we require is using all these metrics together for the selection of the best possible features. In addition to that, since the tree based algorithm get rid of a large chunk of features, it makes computing the recursive algorithms much less computationally expensive. Hence, one may infer that instead of using each method separately before predicting the results, all these methods can be used sequentially so that after each pass, a certain metric would be satisfied. Therefore,  the end result would be those features that pass the tests and are deemed satisfactory.
	
	The final set of features are then passed through a SVC to get the final predictions. In order to choose the best possible parameters for these models, hyper-parameter tuning is performed, both manually and with the help of an algorithm called GridSearchCV, which is an augmentation to our base estimator (SVC) that accepts a bunch of hyper-parameters and returns the set of parameters that are the most optimum for our base estimator model. We also test an alternate method in which Principal Component Analysis (PCA) is used to reduce the dimensions of the datasets and  then SVC is invoked to get the results.
	
	\section{Results and Analyses}\label{sec:results}
	To demonstrate the efficacy of our proposed sequential technique, we train our machine learning model on all the datasets and record the results. As stated earlier, we consider ten datasets, five of which are from RNASeq and the other five are from miRNA expression. The dimensions of the datasets after normalization are presented in \Cref{tab:dimensions}. 
	\begin{table}[!ht]
		\centering
		\caption{Dimensions of data.}
		\label{tab:dimensions}
		\resizebox{\columnwidth}{!}{
			\begin{tabular}{|c|c|c|c|c|}
				\hline
				& \multicolumn{2}{c|}{miRNA} & \multicolumn{2}{c|}{RNASeq} \\ \hline
				& Samples     & Features     & Samples      & Features     \\ \hline
				HNSC & 278         & 412          & 277          & 4466         \\ \hline
				KIRC & 449         & 263          & 443          & 5287         \\ \hline
				KIRP & 160         & 273          & 158          & 5308         \\ \hline
				LUAD & 236         & 433          & 243          & 5112         \\ \hline
				LUSC & 157         & 595          & 179          & 6773         \\ \hline
			\end{tabular}
		}
	\end{table}
	
	We now discuss the results of each test in the coming subsections.	
	
	\subsection{Baseline Test}
\begin{figure}[h!]
		\begin{subfigure}[t]{0.475\columnwidth}
			\centering
			\includegraphics[width=1\linewidth]{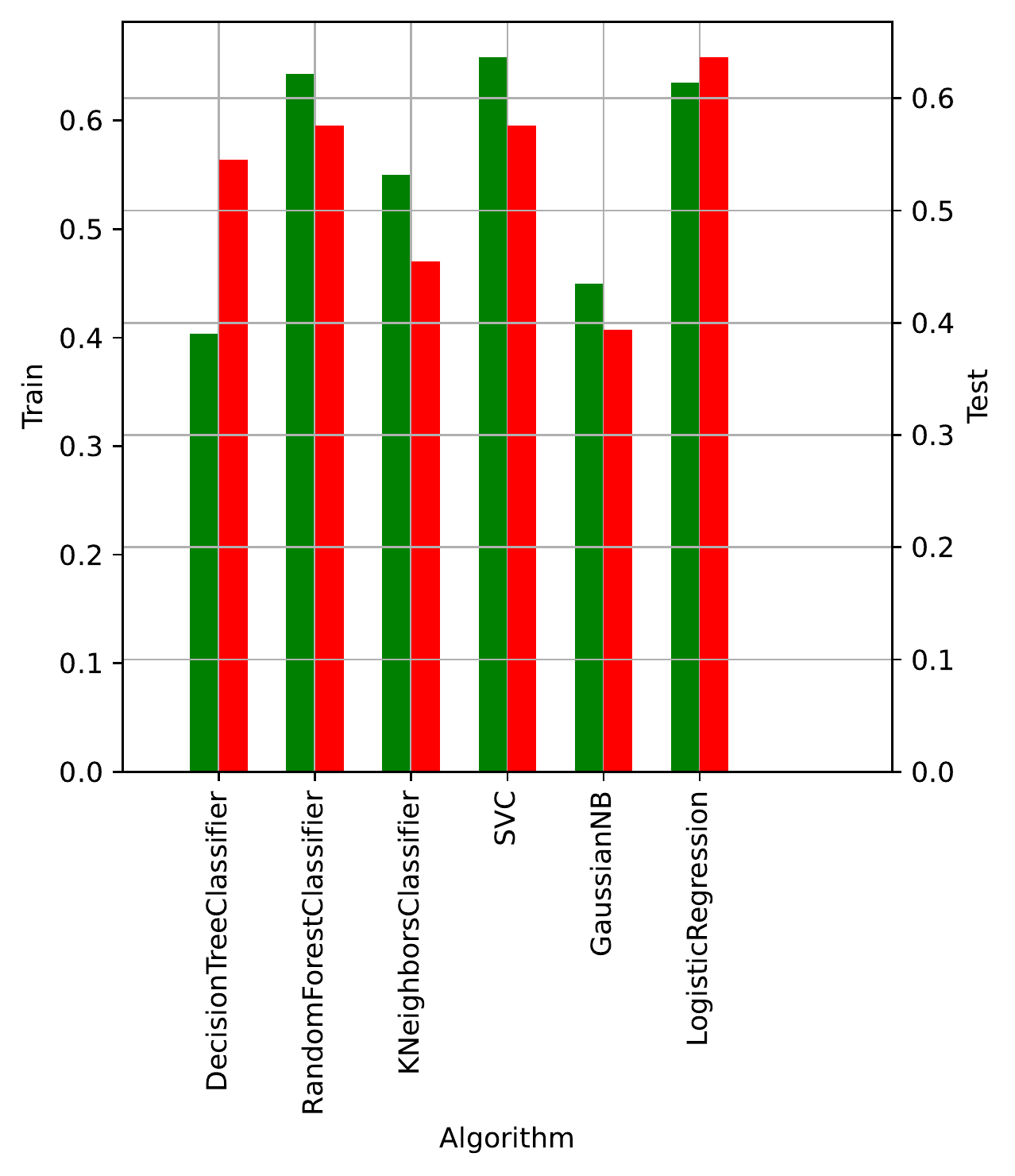}
			\caption{miRNA baseline test.}
			\label{fig:mirna-baseline}
		\end{subfigure}
		\hfill
		\begin{subfigure}[t]{0.475\columnwidth}
			\centering
			\includegraphics[width=1\linewidth]{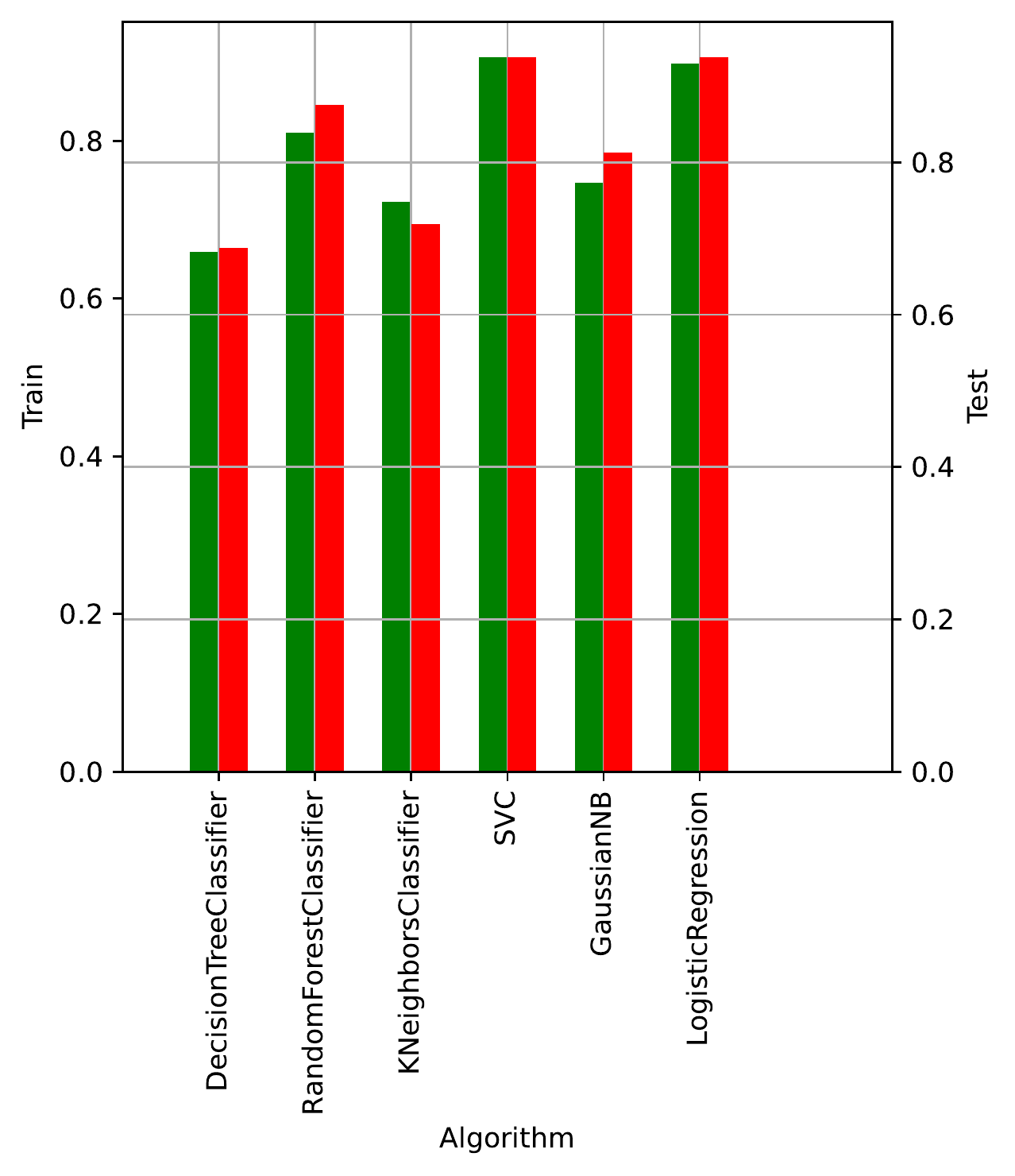}
			\caption{RNASeq baseline test.}
			\label{fig:rnaseq-baseline}
		\end{subfigure}
		\caption{Baseline test results (LUSC).}
		\label{fig:baseline-test-results}
	\end{figure}

	We perform a baseline test on both omics data using six models, namely Decision Tree Classifier, Random Forest Classifier, Logistic Regression, SVC, K Neighbors Classifier, and Gaussian Naive Bayes Classifier. The results of miRNA expression data and RNASeq data are shown in \Cref{fig:mirna-baseline,fig:rnaseq-baseline}, respectively.
	
	One may observe the effect of highly varying, high-dimensional data with low sample size from the results in \Cref{fig:mirna-baseline,fig:rnaseq-baseline}. Due to these problems, the accuracy is either quite low or in some cases the test set has a higher accuracy than the train set. Though not shown here, it readily follows from our analyses that the baseline results on other datasets are also similar. The one model that performs better in comparison to the other five models is the SVC. Thus, we pick SVC to further improve its performance.
	
	\subsection{Feature Selection Results}
	We use three feature selection algorithms, namely, tree-based feature selection, RFECV, and RFE. We use bagged decision trees using the Random Forest estimator in tree-based feature selection. For RFECV and RFE, we use SVC as the estimator, which we selected after the baseline test. As mentioned before, the results are only slightly better than what we get using the baseline models with RFECV producing somewhat better results but taking longer than the other two methods to execute. All three feature selection algorithms cover certain aspects while selecting or eliminating features, but the number of selected features are still quite high.
	\begin{figure}[!ht]
		\begin{subfigure}[t]{0.475\columnwidth}
			\centering
			\includegraphics[width=1\linewidth]{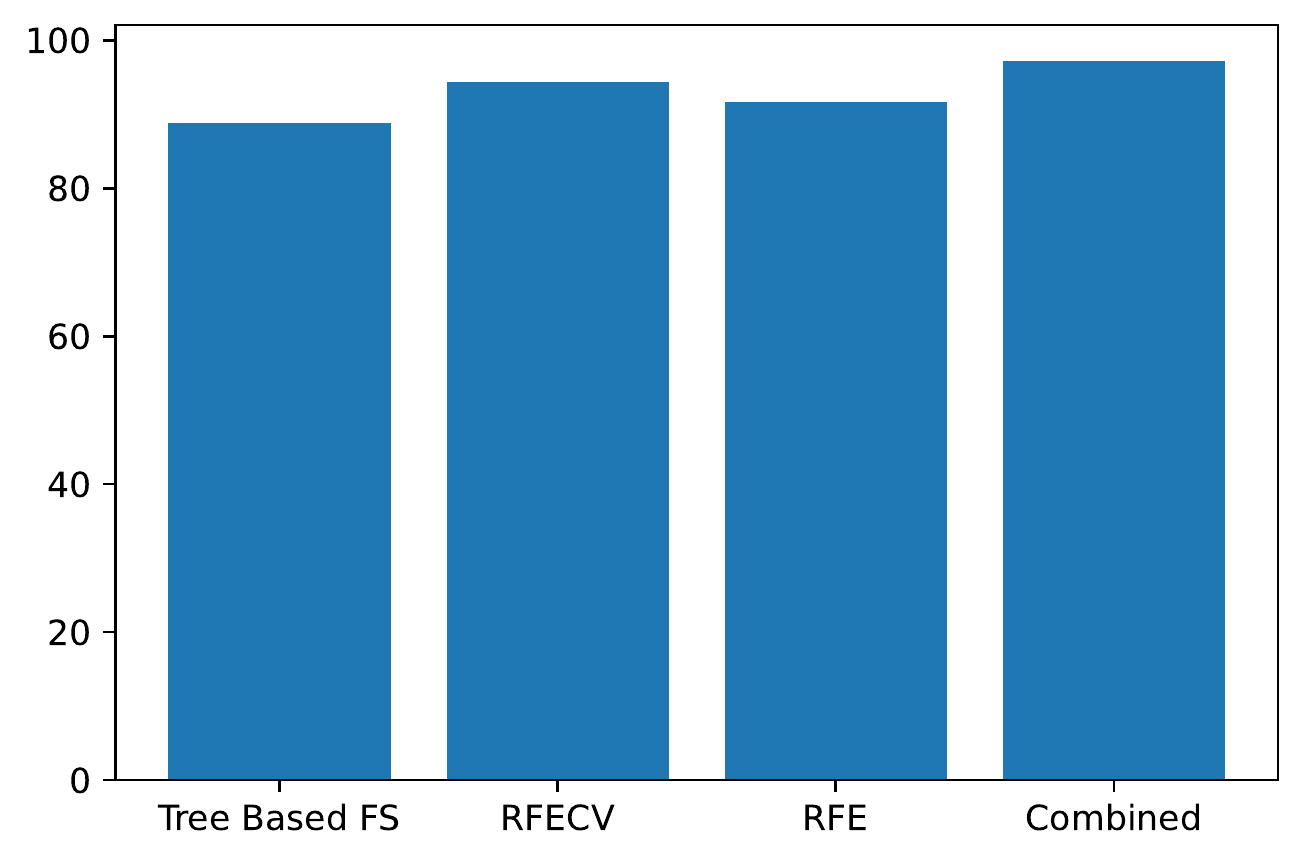}
			\caption{miRNA accuracy.}
			\label{fig:mirna-accuracy}
		\end{subfigure}
		\hfill
		\begin{subfigure}[t]{0.475\columnwidth}
			\centering
			\includegraphics[width=1\linewidth]{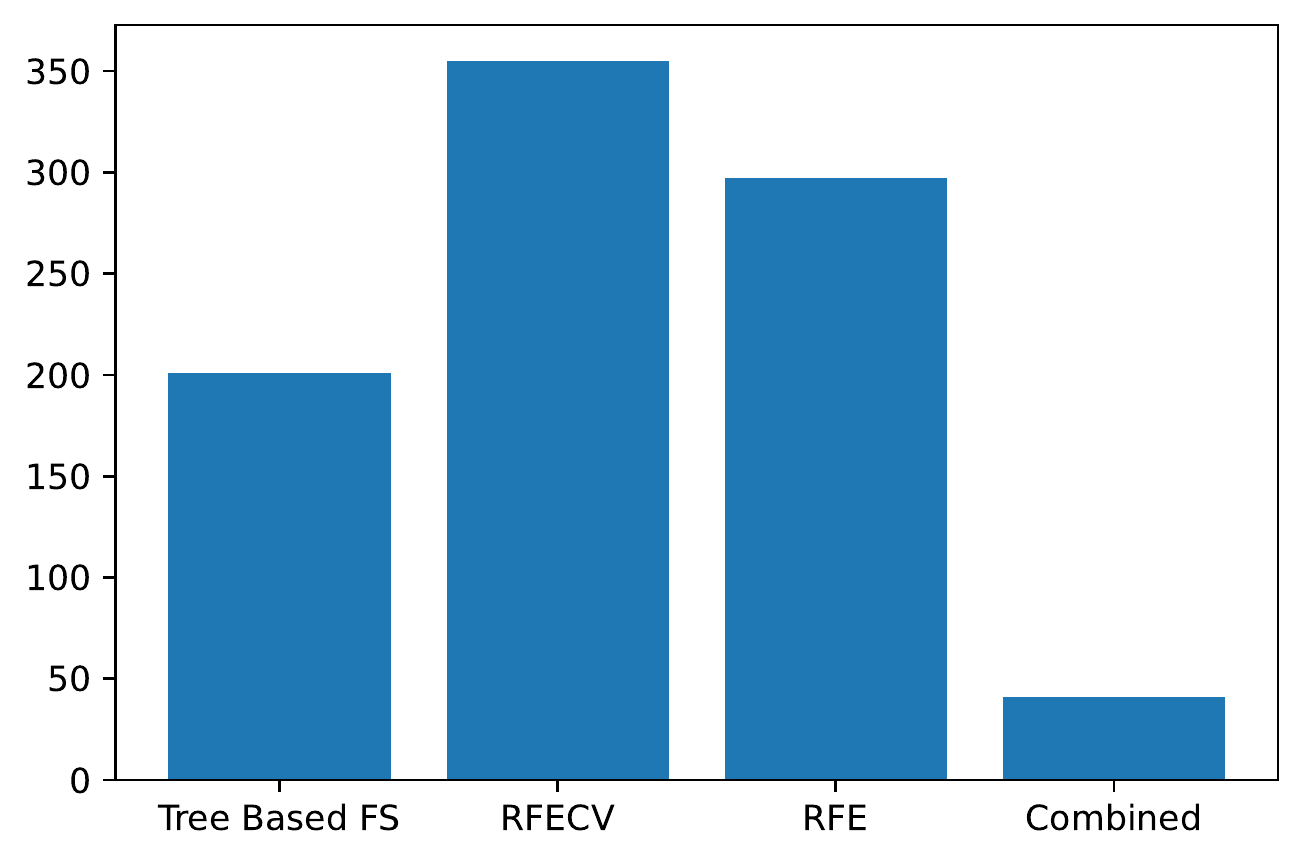}
			\caption{miRNA features.}
			\label{fig:mirna-features}
		\end{subfigure}
		\caption{miRNA Accuracy-Features Relation (LUSC).}
		\label{fig:mirna-accuracy-feature-relation}
	\end{figure}
	
	\begin{figure}[!ht]
		\begin{subfigure}[t]{0.475\columnwidth}
			\centering
			\includegraphics[width=1\linewidth]{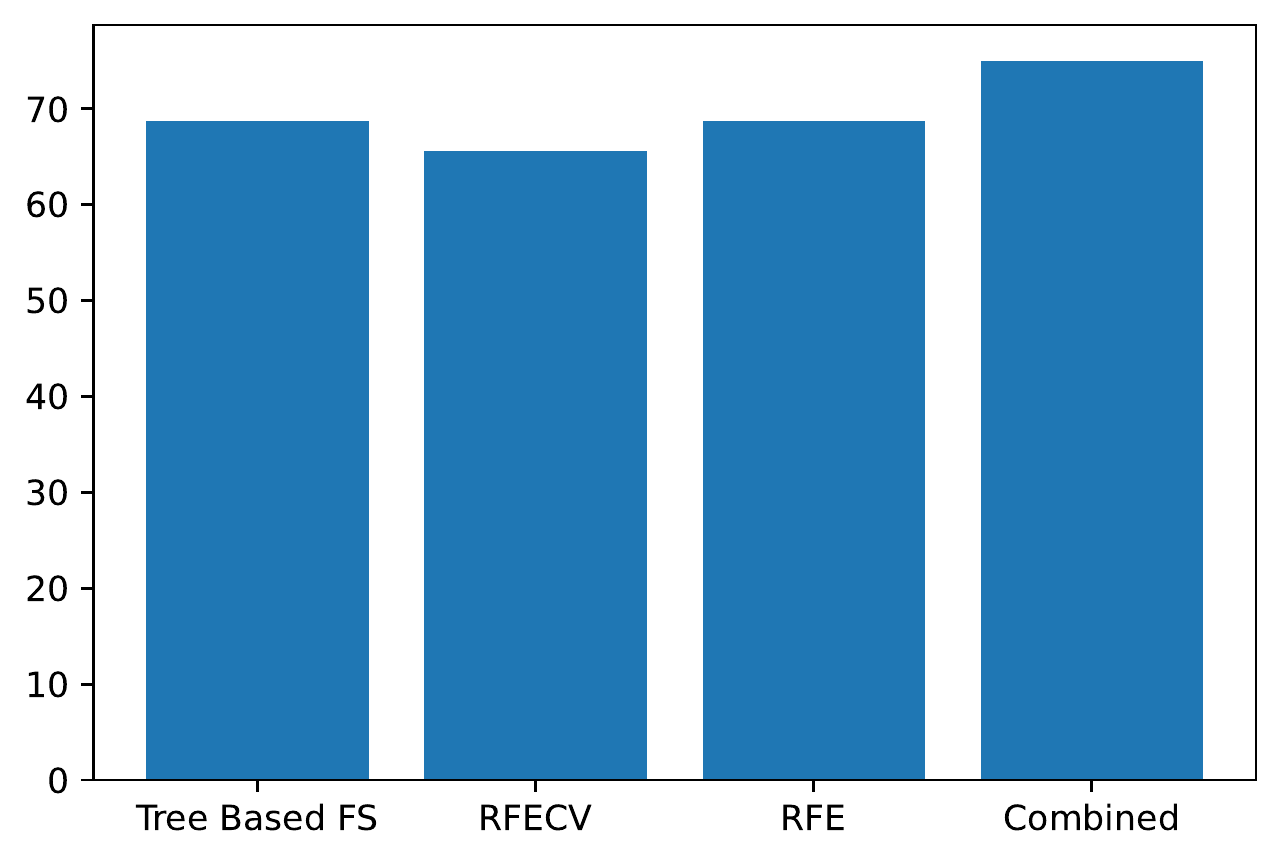}
			\caption{RNASeq accuracy.}
			\label{fig:rnaseq-accuracy}
		\end{subfigure}
		\hfill
		\begin{subfigure}[t]{0.475\columnwidth}
			\centering
			\includegraphics[width=1\linewidth]{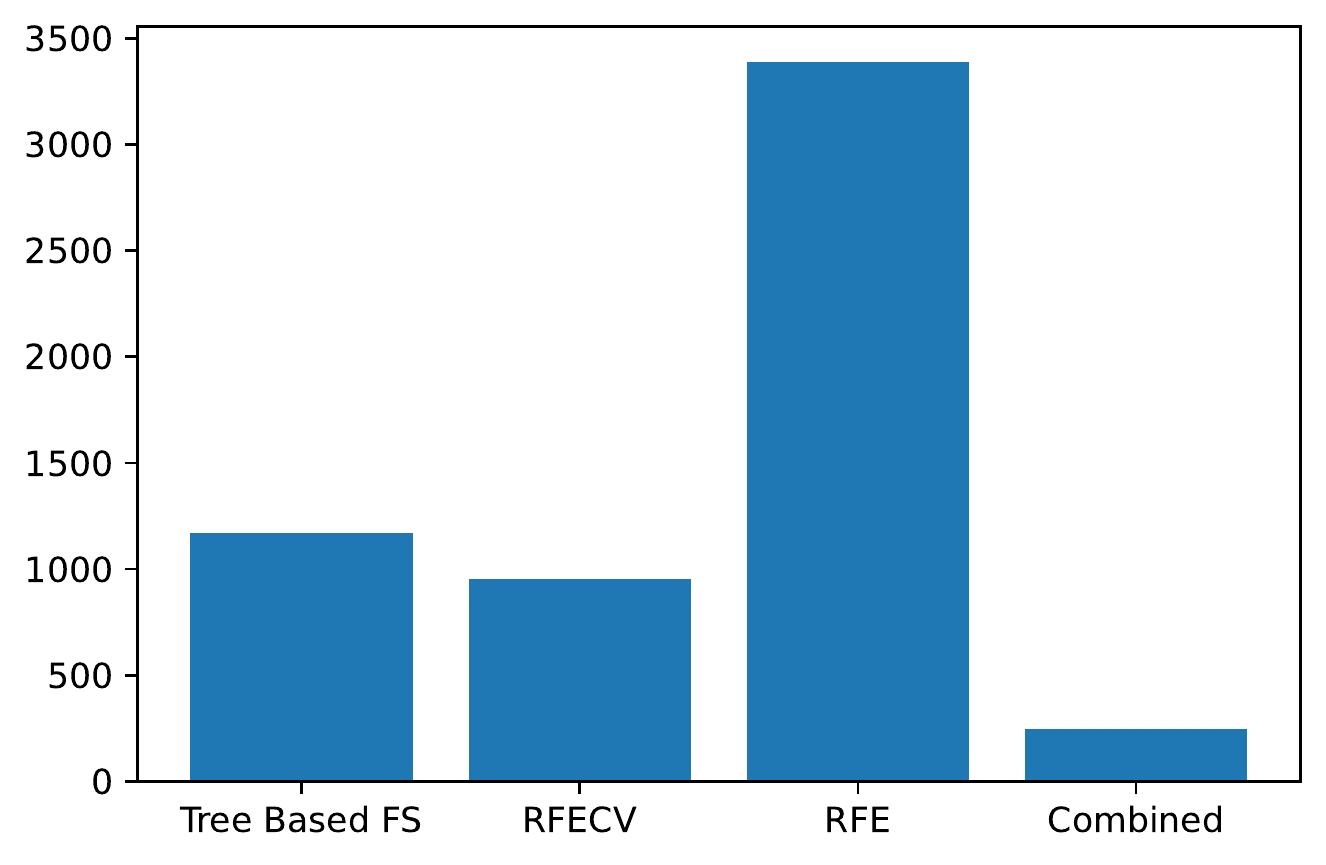}
			\caption{RNASeq features.}
			\label{fig:rnaseq-features}
		\end{subfigure}
		\caption{RNASeq Accuracy-Features Relation (LUSC).}
		\label{fig:rnaseq-accuracy-feature-relation}
	\end{figure}
	
	In order to address the shortcomings associated with the aforementioned techniques, we use these techniques sequentially. We take the features selected from the tree-based feature selection method based on their importance scores and use them to eliminate non-relevant features using RFECV. The stratified k-fold cross-validation in RFECV redces algorithm bias towards some of the features to get the results. Next, we use RFE with the remaining features and apply SVC to make predictions. This improves the overall performance and also lowers the computation cost. A comparative analysis of the results is presented in \Cref{fig:mirna-accuracy-feature-relation,fig:rnaseq-accuracy-feature-relation}.
	
	\begin{table}[!ht]
		\centering
		\caption{Execution time (in seconds).}
		\label{tab:execution-time}
		\begin{tabular}{|c|c|c|}
			\hline
			& RNASeq & miRNA \\ \hline
			Tree-Based & 0.405  & 0.19  \\ \hline
			RFECV      & 212    & 2.94  \\ \hline
			RFE        & 67     & 1.18  \\ \hline
			Combined   & 11.3   & 1.13  \\ \hline
		\end{tabular}%
	\end{table}
	
	The respective execution time of all the feature selection techniques in presented in \Cref{tab:execution-time}. Although tree-based feature selection seems to have the lowest execution time, it also gives the least accuracy when compared to the other methods. Out of the three feature selection techniques, RFECV performs the best, but its execution time is also quite high. However, our proposed method reduces the execution time considerably in addition to improving the accuracy. 
	
	\subsection{Evaluation Metrics}
	Based on the above results and analyses, we proceed with the sequentially applied feature selection model. We present a comparative study of multiple classification metrics in \Cref{tab:rnaseq-metrics,tab:mirna-metrics} to gain additional insights.
	\begin{table}[!ht]
		\centering
		\caption{Evaluation metrics for RNASeq data.}
		\label{tab:rnaseq-metrics}
		\resizebox{\columnwidth}{!}{
			\begin{tabular}{|c|c|c|c|c|c|c|}
				\hline
				& \multicolumn{6}{c|}{Support   Vector Classifier}                \\ \hline
				& Precision & Recall & Specificity & F1 Score & Accuracy & G-Mean \\ \hline
				HNSC & 0.91      & 0.91   & 0.97        & 0.91     & 91.1     & 0.94   \\ \hline
				KIRC & 0.76      & 0.75   & 0.91        & 0.75     & 75.0      & 0.83   \\ \hline
				KIRP & 0.94      & 0.94   & 0.95        & 0.94     & 93.7     & 0.94   \\ \hline
				LUAD & 0.77      & 0.73   & 0.93        & 0.72     & 73.4     & 0.82   \\ \hline
				LUSC & 0.98      & 0.97   & 1.00        & 0.97      & 97.2     & 0.98   \\ \hline
			\end{tabular}
		}
	\end{table}
	\begin{table}[!ht]
		\centering
		\caption{Evaluation metrics for miRNA data.}
		\label{tab:mirna-metrics}
		\resizebox{\columnwidth}{!}{
			\begin{tabular}{|c|c|c|c|c|c|c|}
				\hline
				& \multicolumn{6}{c|}{Support   Vector Classifier}                \\ \hline
				& Precision & Recall & Specificity & F1 Score & Accuracy & G-Mean \\ \hline
				HNSC & 0.79      & 0.77   & 0.93        & 0.77     & 76.8     & 0.84   \\ \hline
				KIRC & 0.73      & 0.71   & 0.88        & 0.70      & 70.8     & 0.78   \\ \hline
				KIRP & 0.76      & 0.75   & 0.75        & 0.74      & 75.7     & 0.73   \\ \hline
				LUAD & 0.72      & 0.71   & 0.94        & 0.70     & 70.8     & 0.80   \\ \hline
				LUSC & 0.77      & 0.75   & 0.89        & 0.73      & 75.0     & 0.80   \\ \hline
			\end{tabular}
		}
	\end{table}
	
	We use Accuracy, Precision, Recall/Sensitivity, Specificity, F1 score, and G-Mean as the metrics to assess the predictive performance of our model.  Accuracy is the total percentage of samples that are correctly predicted. Precision signifies how many samples classified as a particular cancer subtype actually belong to that subtype. Recall (or Sensitivity) signifies how many samples that actually belong to a particular subtype have been predicted as such. Specificity signifies how many samples not classified as a particular cancer subtype actually don't belong to that particular subtype. F1-score is the harmonic mean of precision and recall which balances these metrics. Similarly, G-Mean is the geometric mean of sensitivity and specificity and is used to balance these two metrics. While accuracy can give an overall intuition about how a model is performing, metrics like precision/recall and sensitivity/specificity can better evaluate the usefulness of the model.
	
	As presented in \Cref{tab:rnaseq-metrics,tab:mirna-metrics},  RNASeq datasets have very high values of specificity and quite high values of sensitivity as well. On the other hand, miRNA datasets have relatively low sensitivity but high specificity. High sensitivity means our model identifies maximum number of patients suffering from a particular cancer subtype correctly. High specificity means our model can be used to cross-check if a patient identified as not having a particular cancer subtype is correct. There is usually a trade-off between precision and recall, and sensitivity and specificity. Tuning the model to increase one of these metrics decreases the other, and vice-versa. Thus, F1-score and G-mean are used to evaluate the model properly.
	
	\subsection{Visual Analyses}
\begin{figure}[!ht]
		\begin{subfigure}[t]{0.475\columnwidth}
			\centering
			\includegraphics[width=1\linewidth]{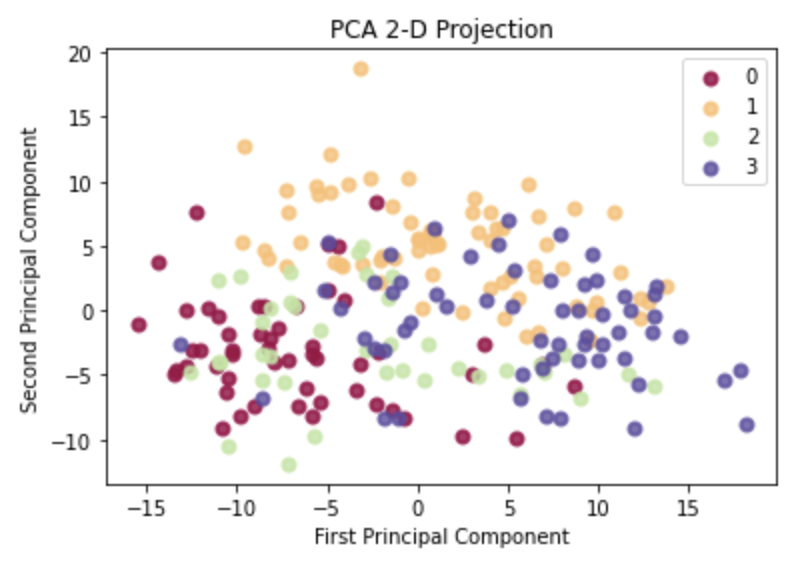}
			\caption{miRNA PCA plot.}
			\label{fig:mirna-pca}
		\end{subfigure}
		\hfill
		\begin{subfigure}[t]{0.475\columnwidth}
			\centering
			\includegraphics[width=1\linewidth]{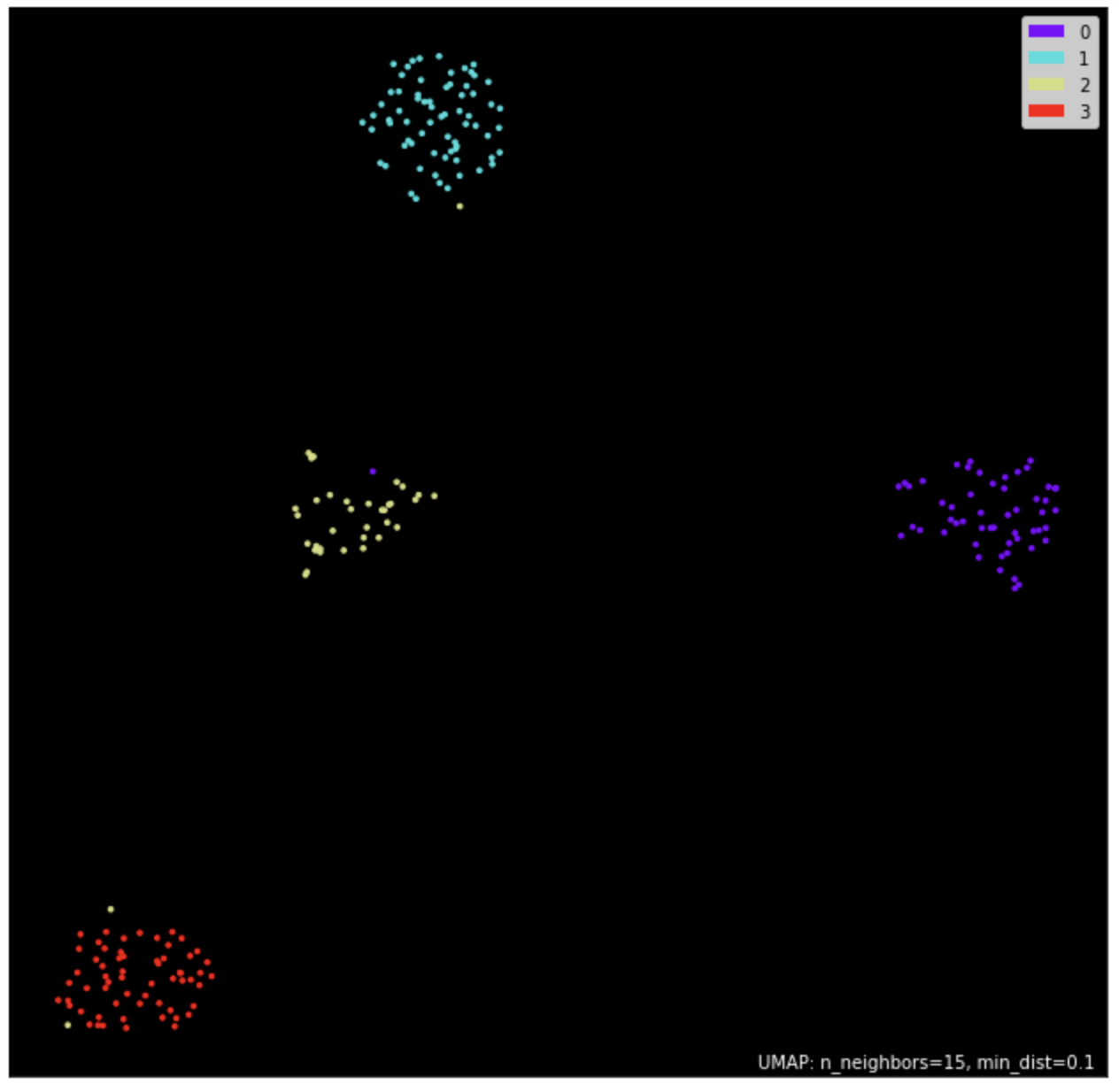}
			\caption{miRNA UMAP plot.}
			\label{fig:mirna-umap}
		\end{subfigure}
		\caption{PCA versus UMAP.}
		\label{fig:pca-umap}
	\end{figure}

	We perform techniques like PCA and UMAP \cite{mcinnes2018umap-software} on the data to gain some visual insights regarding the difference between the performance of both miRNA and RNASeq datasets. UMAP clustered the data into different classes comparatively better than PCA. A comparision of PCA and UMAP is presented in \Cref{fig:pca-umap}.
	\begin{figure}[!ht]
		\begin{subfigure}[t]{0.475\columnwidth}
			\centering
			\includegraphics[width=1\linewidth]{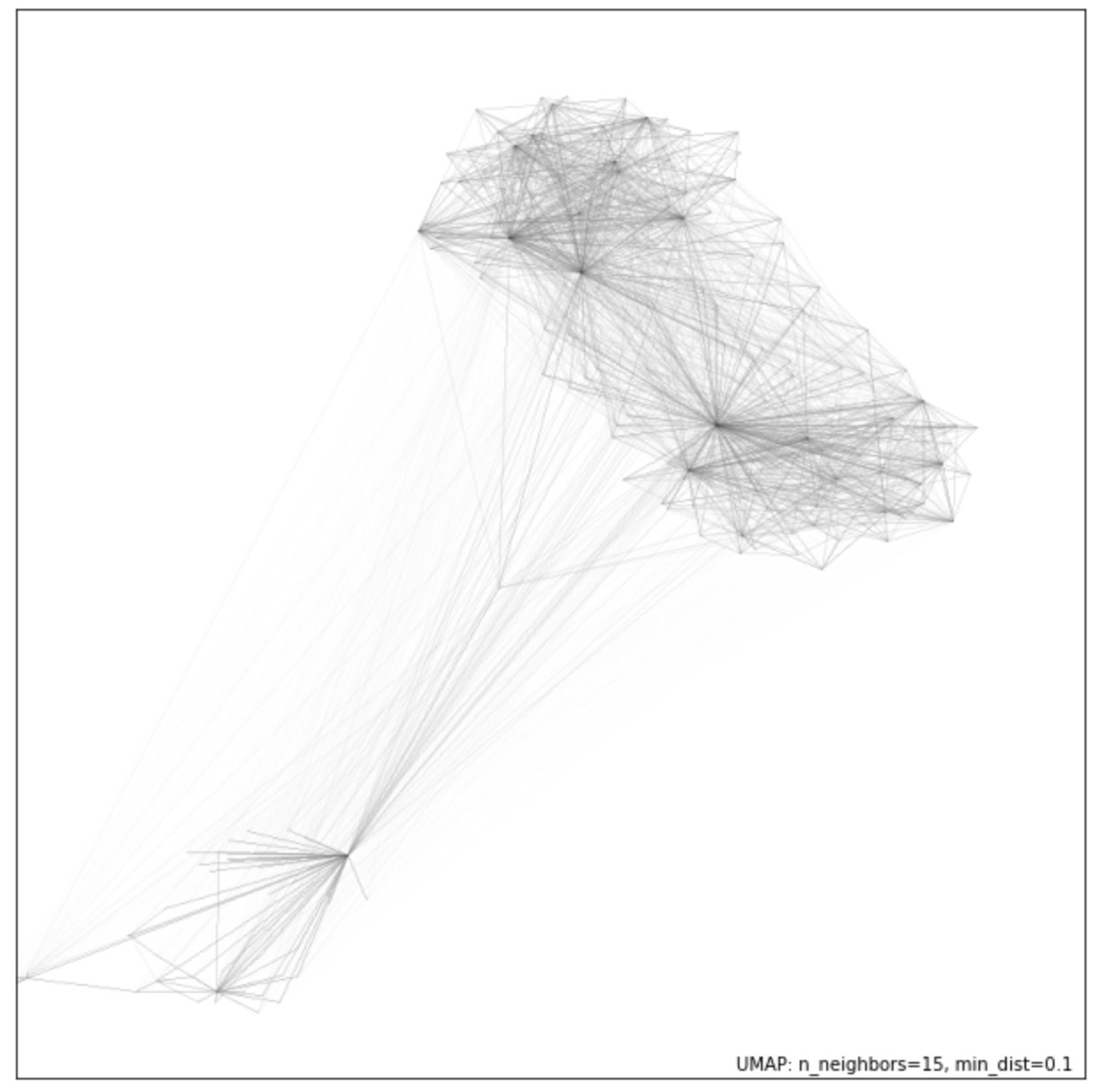}
			\caption{RNASeq connectivity graph.}
			\label{fig:hnsc-rnaseq-connectivity}
		\end{subfigure}
		\hfill
		\begin{subfigure}[t]{0.475\columnwidth}
			\centering
			\includegraphics[width=1\linewidth]{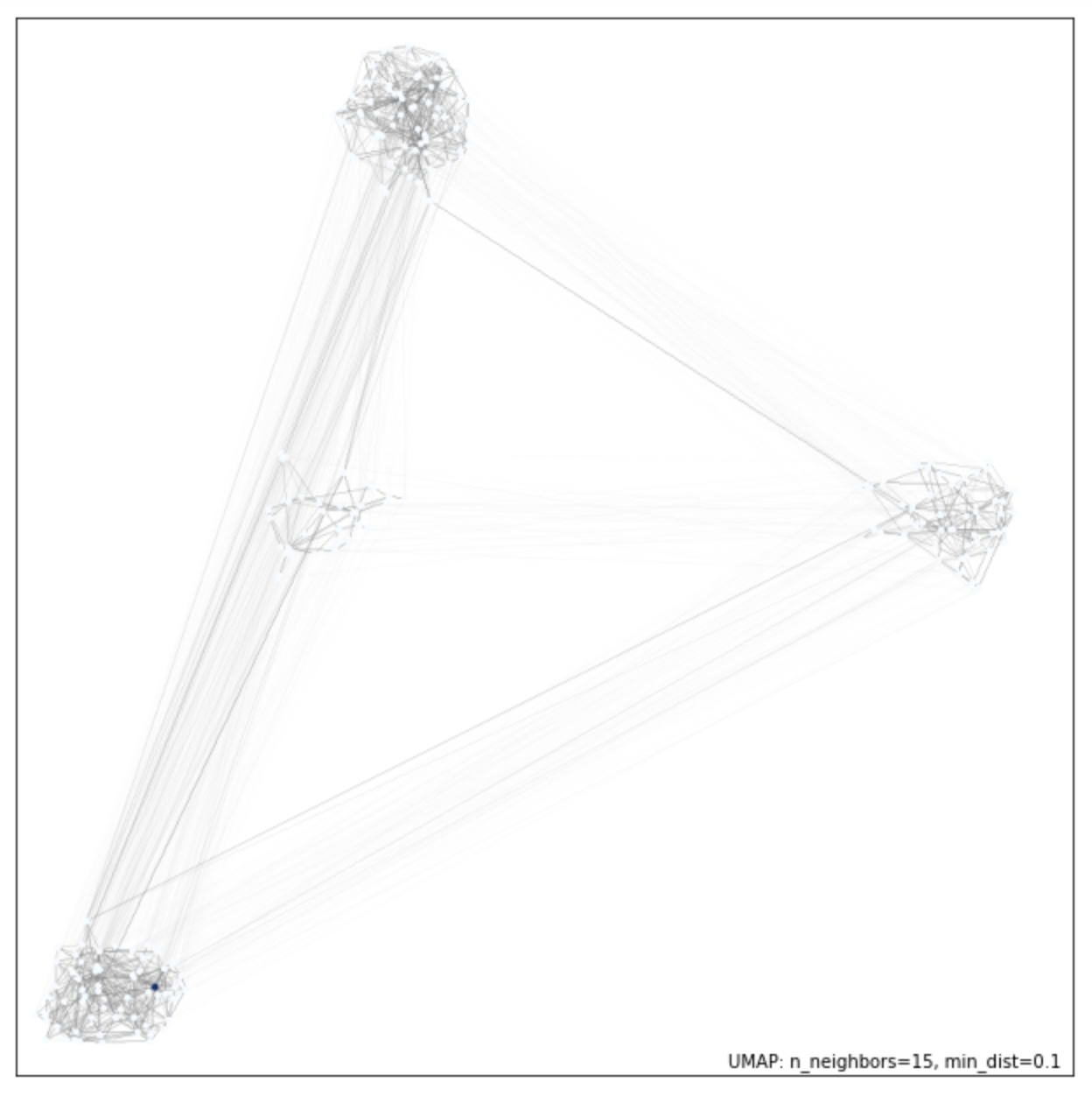}
			\caption{miRNA connectivity graph.}
			\label{fig:hnsc-mirna-connectivity}
		\end{subfigure}
		\caption{HNSC connectivity graph.}
		\label{fig:hnsc-connectivity}
	\end{figure}
	\begin{figure}[!ht]
		\begin{subfigure}[t]{0.475\columnwidth}
			\centering
			\includegraphics[width=1\linewidth]{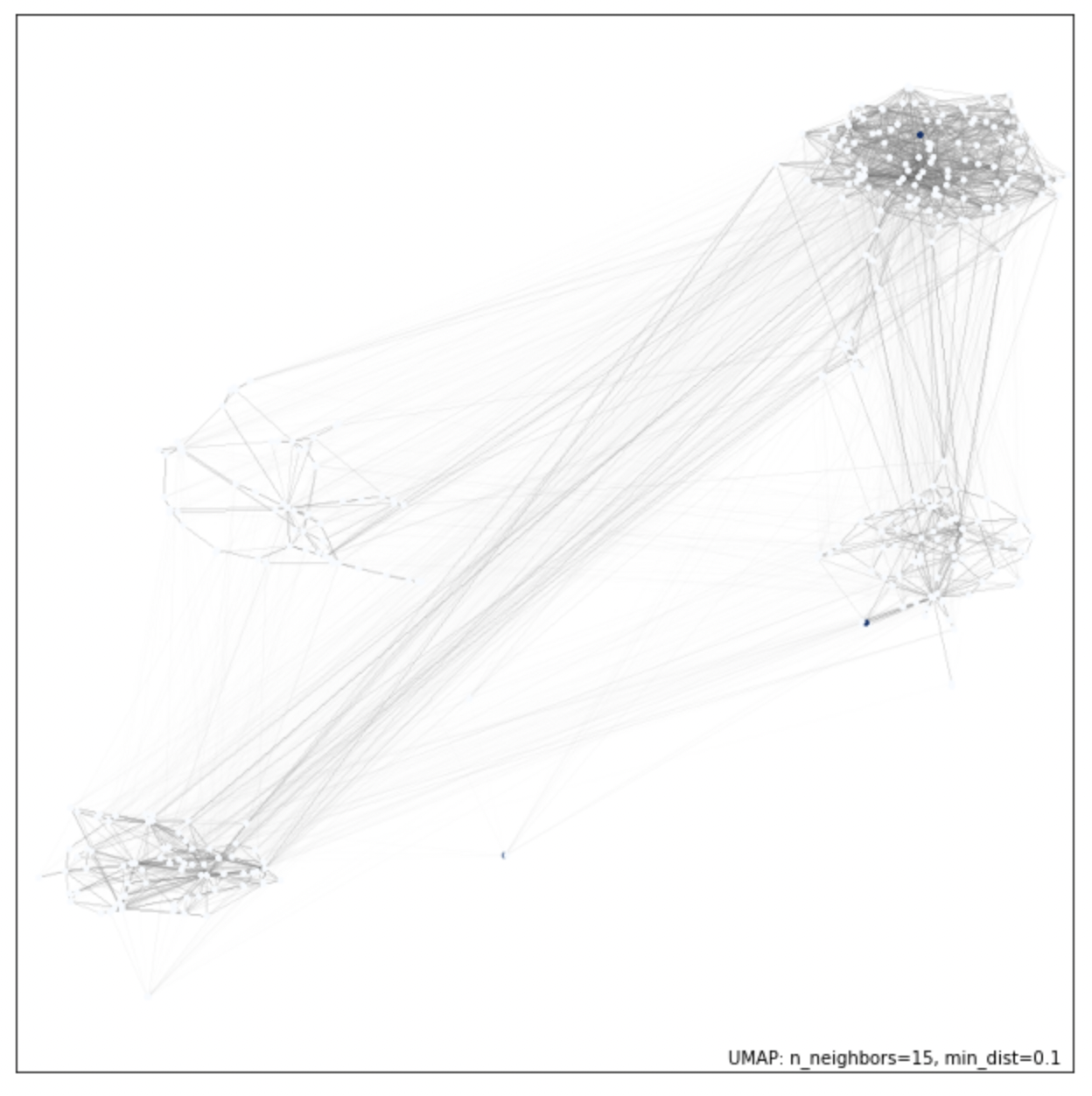}
			\caption{RNASeq connectivity graph.}
			\label{fig:kirc-rnaseq-connectivity}
		\end{subfigure}
		\hfill
		\begin{subfigure}[t]{0.475\columnwidth}
			\centering
			\includegraphics[width=1\linewidth]{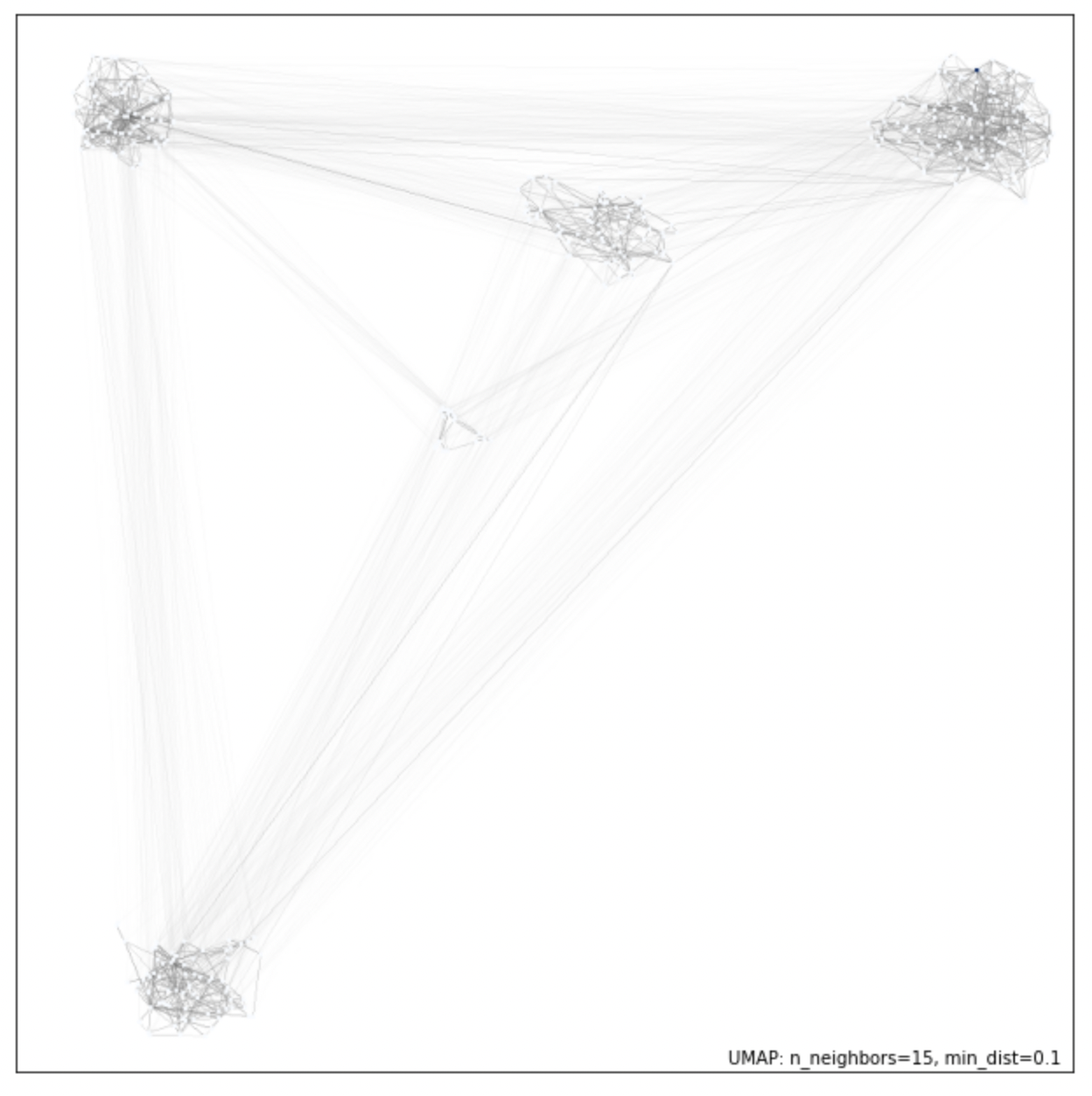}
			\caption{miRNA connectivity graph.}
			\label{fig:kirc-mirna-connectivity}
		\end{subfigure}
		\caption{KIRC connectivity graph.}
		\label{fig:kirc-connectivity}
	\end{figure}
	
	We use the result from UMAP to plot weighted graphs which are the intermediate topological representation of the approximate manifold the data has been sampled from. Intuitively, it gives us an idea of how interrelated the data points in the dataset are, both within the clusters and between the clusters. RNASeq showed greater connectivity in the dataset as compared to corresponding miRNA dataset which is depicted in \Cref{fig:hnsc-connectivity,fig:kirc-connectivity}. Visually, this can be inferred by the thickness of the weighted components of the graphs plotted. This connectivity in the manifold shows why the RNASeq datasets outperform the corresponding miRNA datasets.

	\section{Conclusions}\label{sec:conclusions}
	We studied the transcriptomic data of five different cancer types using the miRNA and RNASeq expression values, and it was found that a combination of multiple feature selection algorithms helps in reducing the computational expense in addition to improving the accuracy of the prediction. Our model was highly specific in case of miRNA data but maintained a good balance between specificity and sensitivity in case of RNASeq data. Therefore, the model can be used for the final confirmation of the disease when used with miRNA data, and for initial diagnois as well as final confirmation of disease when used with RNASeq data.  The results in this work provided a strong foundation in cancer subtype classification. One such area is the classification of cancer subtypes using integrated multi-omics data which may be an interesting investigation for future studies.
	
	\bibliographystyle{IEEEtran}
	\bibliography{references}
\end{document}